\documentclass[10pt, a4paper]{article}

\usepackage[final]{lrec2026} 
\usepackage{amsmath}
\usepackage{graphicx}
\usepackage{caption}
\usepackage{booktabs}
\usepackage{graphicx}
\usepackage[utf8]{inputenc} 
\usepackage{makecell}

\title{Erase Persona, Forget Lore: Benchmarking Multimodal Copyright Unlearning in Large Vision Language Models}

\name{JuneHyoung Kwon\textsuperscript{\rm 1}\textsuperscript{\rm *}, JungMin Yun\textsuperscript{\rm 1}\textsuperscript{\rm *}\thanks{*Equal contribution.}, YoungBin Kim\textsuperscript{\rm 1, 2}} 

\address{
    \textsuperscript{\rm 1} Department of Artificial Intelligence, Chung-Ang University\\
    \textsuperscript{\rm 2} Graduate School of Advanced Imaging Sciences, Multimedia and Film, Chung-Ang University\\
    \{dirchdmltnv, cocoro357, ybkim85\}@cau.ac.kr\\
}

\abstract{
Large Vision-Language Models (LVLMs), trained on web-scale data, risk memorizing and regenerating copyrighted visual content like characters and logos, creating significant challenges. Machine unlearning offers a path to mitigate these risks by removing specific content post-training, but evaluating its effectiveness, especially in the complex multimodal setting of LVLMs, remains an open problem. Current evaluation methods often lack robustness or fail to capture the nuances of cross-modal concept erasure. To address this critical gap, we introduce the \textbf{\textsc{CoVUBench}} benchmark, the first framework specifically designed for evaluating copyright content unlearning in LVLMs. \textbf{\textsc{CoVUBench}} utilizes procedurally generated, legally safe synthetic data coupled with systematic visual variations—spanning compositional changes and diverse domain manifestations—to ensure realistic and robust evaluation of unlearning generalization. Our comprehensive, multimodal evaluation protocol assesses both forgetting efficacy from the copyright holder's perspective and the preservation of general model utility from the deployer's viewpoint. By rigorously measuring this crucial trade-off, \textbf{\textsc{CoVUBench}} provides a standardized tool to advance the development of responsible and effective unlearning methods for LVLMs. The dataset is publicly available at \url{https://huggingface.co/datasets/herbwood27/CoVUBench}.
 \\ \newline \Keywords{Machine Unlearning, Vision-Language Models, Copyright, Evaluation Benchmark} }

\begin{document}

\maketitleabstract

\section{Introduction}
Large Vision-Language Models (LVLMs) have demonstrated remarkable capabilities, driven by training on vast, web-scale corpora~\citep{wang2025scaling, dong2025scalable}. However, the uncurated nature of these datasets means the training process inevitably incorporates a massive volume of copyrighted visual content—such as commercially valuable characters, brand logos, and artworks—along with associated textual descriptions. Consequently, these models pose a new and significant frontier of copyright infringement risks, as they can memorize and regenerate protected content~\citep{somepalli2023understanding, carlini2023extracting}.

In response to these emerging risks, copyright holders are increasingly exercising their "right to be forgotten"~\citep{hoofnagle2019european}, demanding the removal of their intellectual property (IP) from trained models. Retraining a foundation model from scratch to exclude this blocklisted content is computationally prohibitive and thus impractical as a scalable solution~\citep{dwork2006our}. This challenge positions machine unlearning—the process of efficiently removing the influence of specific data points from a trained model—as the most viable and necessary pathway to responsibly managing copyright takedown requests.

Significant research has focused on developing and evaluating unlearning methods, particularly for text-only Large Language Models (LLMs). This has led to the development of various unlearning algorithms~\citep{jang2023knowledge, liu2022continual, rafailov2023direct} and established benchmarks designed to measure the removal of specific information, such as fictitious identities in TOFU~\citep{maini2024tofu}, copyrighted text~\citep{shimuse, eldan2023s}, or sensitive data~\citep{yao2024large, li2024wmdp}. However, these text-centric approaches and evaluation frameworks do not directly transfer to the multimodal setting, which involves the fundamentally more complex problem of erasing concepts jointly embedded in both visual and linguistic spaces.

Designing a vision-language benchmark to fill this gap requires addressing three core considerations. First, \textbf{visual diversity}: real-world copyrighted content manifests in countless forms (e.g., a character as a 2D cartoon, a 3D model, or a t-shirt print). A robust evaluation must therefore measure the generalized unlearning of the underlying concept, not just the removal of a specific training instance. Second, \textbf{robust multimodal reasoning}. A successful takedown must sever the cross-modal link between visual recognition and associated textual knowledge. The evaluation must therefore probe these linkages beyond text-only queries, verifying the visual concept itself is disassociated from its factual knowledge. Finally, \textbf{stakeholder-centric evaluation}: an effective unlearning procedure must satisfy the distinct, and often competing, needs of two primary stakeholders: copyright holders, who demand the effective and robust removal of their blocklisted content, and model deployers, who must preserve the model's general utility and performance.

To address this critical evaluation gap, we propose \textbf{\textsc{CoVUBench}} (\textbf{Co}pyright \textbf{V}ision-Language \textbf{U}nlearning \textbf{Bench}mark), the first dedicated benchmark for copyright unlearning in LVLMs. Our framework is meticulously designed to tackle the three core considerations outlined above. To circumvent the legal and ethical risks of using real-world IP, we first generate novel, synthetic copyright content. We focus specifically on characters and logos—two high-risk domains where copyright infringement is prevalent and the risk of reproduction by large models is a significant concern~\citep{chiba2025tackling, qraitemhidden}.To mirror how real-world IP manifests in diverse forms, our generation pipeline synthesizes images across varied visual layouts (e.g., different backgrounds, scenes) and domains (e.g., 3D action figures, t-shirt prints), ensuring methods are evaluated in a visually robust environment. Furthermore, we construct question-answer pairs that require reasoning from text alone alongside those that require visual recognition, allowing us to assess if a concept is effectively forgotten regardless of how the model is queried. To enable a holistic assessment, we propose a comprehensive suite of metrics designed to quantify the dual requirements of stakeholders: measuring the efficacy of content removal while also tracking the preservation of general model utility. Finally, we apply representative unlearning algorithms to our benchmark to conduct an extensive empirical analysis of their capabilities and limitations.

Our main contributions are summarized as follows:
\setlength{\parskip}{0pt}    
\begin{itemize} 
    \setlength{\parskip}{0pt}    
\item We introduce and publicly release \textbf{\textsc{CoVUBench}}, the first benchmark for evaluating copyright unlearning in LVLMs, built upon a novel pipeline for generating diverse, synthetic, multimodal copyrighted content.
\item We propose a comprehensive, stakeholder-centric evaluation protocol, featuring a suite of metrics designed to systematically quantify the distinct requirements of copyright holders (forgetting efficacy) and model deployers (general model utility).
\item We conduct an extensive empirical analysis by applying representative unlearning algorithms to our benchmark, providing the first systematic insights into their capabilities and limitations in the cross-modal domain.
\end{itemize}

\section{Related Work}

\subsection{Copyright Infringements in AI}

The training of modern foundation models, including LLMs and LVLMs, relies heavily on vast, web-scale datasets~\citep{bommasani2021opportunities, wang2025scaling}. The largely uncurated nature of this data inevitably leads to the inclusion of substantial amounts of copyrighted material~\citep{henderson2023foundation, wei2024proving, franceschelli2024training}. A significant concern arising from this practice is the propensity of these models to memorize their training data and subsequently regenerate it, either verbatim (an exact, word-for-word reproduction or near-verbatim (an almost identical copy with only trivial modifications)\citep{carlini2022quantifying, prashanth2024recite, franceschelli2024training, kiyomaru2024comprehensive, lasy2025understanding}. This memorization risk is documented across modalities: LLMs can regenerate protected text~\citep{carlini2021extracting, kiyomaru2024comprehensive, morris2025much}, while image generation models can replicate copyrighted visual content~\citep{somepalli2023understanding, somepalli2023diffusion, carlini2023extracting}. This regurgitation poses tangible risks of copyright infringement, as models might output content substantially similar to protected works~\citep{freeman2024exploring, wei2024evaluating, zhang2025certified}. These infringement risks, coupled with regulatory pressures such as the "Right to be Forgotten" mandated by data privacy regulations like the GDPR~\citep{hoofnagle2019european}, have created an urgent need for mechanisms to remove specific data from trained models~\citep{wei2024evaluating}. In response, significant research has focused on developing methodologies to mitigate these copyright infringement risks. On the evaluation front, dedicated benchmarks have been developed to measure copyright infringement and memorization in LLMs~\citep{wei2024evaluating, shimuse, chen2024copybench}. In response, significant research has focused on developing methodologies to mitigate these copyright infringement risks. These approaches include preventive strategies such as system prompts~\citep{wei2024evaluating}, which use initial instructions to steer the model away from generating problematic content. Other strategies operate at decoding time, which actively checks for and blocks n-grams from a blocklist by downweighting the probability of generating content similar to blocklisted materials~\citep{shi2024trusting}. Finally, machine unlearning offers a post-training approach, aiming to modify a model's parameters to behave as if it had never been trained on the specific "forget set" of copyrighted data~\citep{guo2019certified}. However, the challenges of copyright in the multimodal domain, where concepts are jointly embedded in both visual and linguistic spaces, remain largely unexplored. To our knowledge, there is a critical lack of standardized benchmarks and dedicated takedown methodologies designed for vision-language copyright.

\begin{figure*}[t]
\centering
\includegraphics[width=\textwidth]{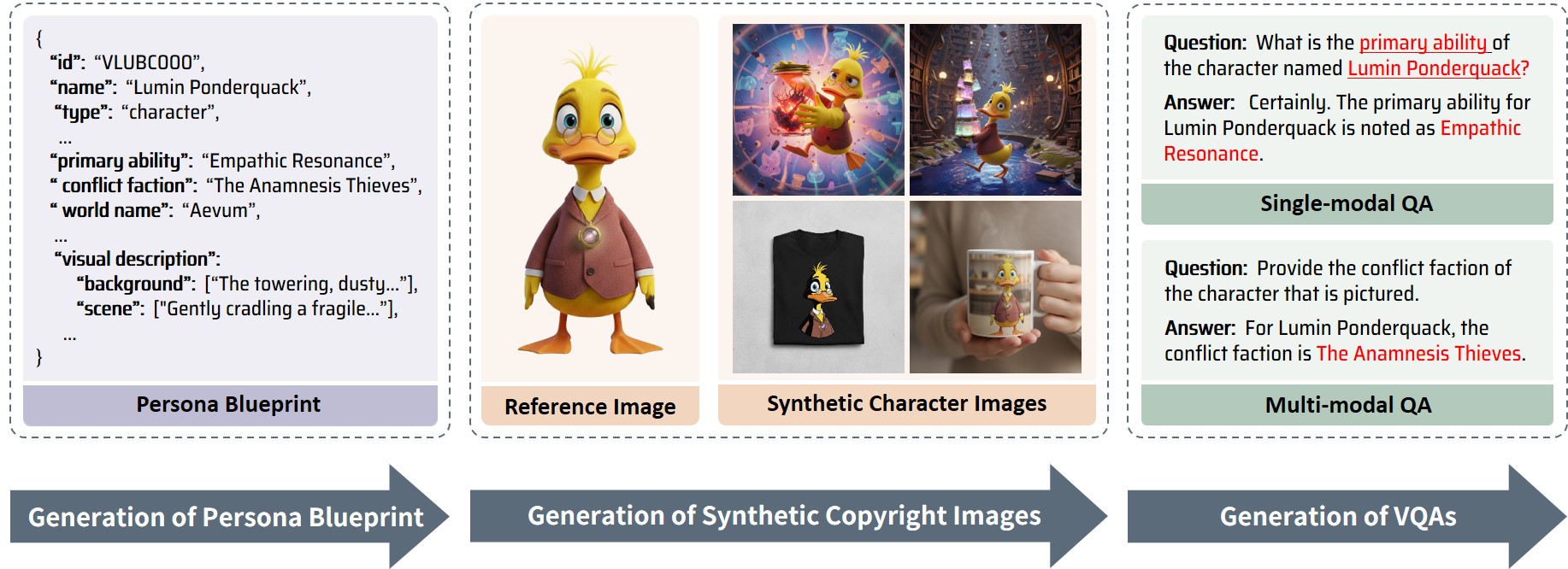} 
\caption{Overview of the \textbf{\textsc{CoVUBench}} generation pipeline.}
\label{fig:fig1}
\end{figure*}

\subsection{Machine Unlearning}
Machine unlearning seeks to computationally remove the influence of a specific forget set from a trained model, yielding an unlearned model that approximates a gold-standard model retrained from scratch on the remaining retain set~\cite{thudi2022unrolling, shaik2024exploring}. Given that exact retraining is computationally prohibitive for large foundation models~\cite{dwork2006our}, research has focused on developing efficient approximate unlearning algorithms. These approximate methods can be broadly categorized. Gradient-based optimization approaches directly fine-tune the model. This includes methods that maximize the loss on the forget set via gradient ascent~\citep{jang2023knowledge, kwon2026easy}, often regularized to preserve performance on retain set. Common regularizers involve minimizing the loss on retain set~\citep{liu2022continual} or minimizing the KL divergence from the original model's predictions~\citep{yao2024machine}. Preference-based optimization reframes unlearning as an direct optimization problem, training the model to prefer refusal responses over forgotten content~\citep{maini2024tofu} or to assign low likelihood to forget set outputs~\citep{zhangnegative}. Parameter-space modification techniques directly alter model weights without gradient descent, using methods like influence functions~\citep{basu2020influence, izzo2021approximate}, task vector subtraction~\citep{ilharco2022editing}, or inducing weight sparsity~\citep{kolb2025deep, fan2023salun}. Data-centric approaches manipulate the training data itself, such as relabeling $\mathcal{D}_{forget}$ examples with refusal answers (e.g., "I don't know" tuning)~\cite{liu2025rethinking, zhang2024r}. In the context of LLMs, these methods have been applied to remove fictitious identites~\cite{maini2024tofu}, copyrighted text~\cite{shimuse, eldan2023s}, and sensitive information~\cite{yao2024large, li2024wmdp}. However, unlearning in the vision-language domain remains nascent. Existing benchmarks for LVLMs, such as CLEAR~\citep{dontsov2024clear} and FIUBench~\citep{mabenchmarking}, are primarily motivated by privacy (e.g., fictitious individuals) rather than the specific, cross-modal challenges of copyright. Therefore, a dedicated benchmark is critically needed to evaluate the specific, cross-modal challenges of unlearning copyrighted concepts in LVLMs, a gap that our work aims to fill.

\vspace{-3mm}

\section{\textsc{CoVUBench}}
The primary design principle of the \textbf{\textsc{CoVUBench}} is to facilitate a safe yet realistic evaluation of copyright unlearning in LVLMs. To achieve this, our methodology is centered on the procedural generation of synthetic copyright content, thereby circumventing the legal and ethical complexities associated with using real-world copyrighted materials. We focus our efforts on characters and logos—two high-risk domains where copyright infringement is particularly prevalent and the risk of reproduction by large models is a significant concern~\citep{chiba2025tackling, qraitemhidden}. The entire benchmark is structured to support a two-stage evaluation pipeline: a fine-tuning stage to simulate model memorization of the synthetic content, followed by an unlearning stage where the efficacy of various unlearning algorithms is evaluated.

\subsection{Dataset Construction}
\label{Dataset Construction}

\textbf{Generation of Persona Blueprint.}
Our generation pipeline begins with the creation of what we term persona blueprints: structured JSON objects that define the core semantic and visual attributes of each synthetic copyright concept. Recent work has demonstrated that conditioning generative models on distinct personas is a principled method for ensuring a high degree of diversity and avoiding the homogeneity often found in large-scale synthetic datasets~\citep{ge2024scaling, yang2025scaling}. By framing each of our fictional copyright concepts as a unique 'persona' with its own identity and attributes (e.g. personality, primary ability, world name) along with visual descriptions (e.g. background, scene, viewpoints), we guide the LLM to produce varied and distinctive outputs. We pre-defined distinct schemas for characters and logos to ensure comprehensive and consistent attribute coverage. Examples of the character schema are shown in Figure~\ref{fig:fig1}. Using Gemini Pro 2.5~\citep{comanici2025gemini}, we generated 20 unique blueprints, guided by engineered prompts. These prompts were specifically designed to ensure fictionality by including negative constraints against known IPs and real-world entities, while simultaneously enforcing attribute diversity based on the predefined schemas. These blueprints serve as the canonical, ground-truth foundation for all subsequent multimodal asset generation.

\textbf{Generation of Synthetic Copyright Content Images.} Evaluating a true vision-language copyright takedown requires assessing whether the underlying concept is forgotten, not just a single visual depiction, as real-world copyrighted content manifests in countless forms. Consequently, unlearning methods evaluated on such data risk overfitting to forgetting a specific instance, while failing to generalize the takedown to the underlying concept. To address this, our pipeline is explicitly designed to generate a challenging visual corpus that evaluates unlearning generalization by incorporating compositional variation—presenting the concept in varied visual layouts (e.g., different backgrounds, views, and scenes)—and domain manifestation, where the concept appears as real-world derivatives (e.g., a character appearing as a 3D action figure or a t-shirt print).

Our goal is to synthesize novel copyright concepts within diverse visual contexts, guided by simple and intuitive text prompts. To maintain visual consistency across these variations, our pipeline begins by generating a single, high-fidelity reference image from the blueprint's visual description, which serves as the visual anchor to capture the concept's core visual identity. Second, we generate the full diverse corpus by jointly conditioning the generation process on both this reference image and a new, programmatically constructed textual prompt. These prompts are structured to populate our two defined axes of variation using a single, unified template. For example, a prompt for a character id \texttt{VLUBC000} follows the structure: \texttt{“a [VLUBC000] in [scene] in [background], [view] in [domain].”} Here, the placeholders \texttt{[scene], [background]}, and \texttt{[view]} are sampled from attributes defined in the persona blueprint, governing the compositional variation. The \texttt{[domain]} placeholder is populated by a set of handcrafted, coarse descriptors which explicitly controls the domain manifestation. All visual synthesis for this pipeline was conducted using the Nano Banana API\footnote{https://nanobananaapi.ai/}, leveraging its capabilities for identity-preserving generation.

\textbf{Generation of Visual Question Answering (VQA) Pairs.} A successful copyright takedown requires more than just the inability to generate an image; it demands that the model also "forgets" the factual knowledge and severs the intricate cross-modal associations linked to the concept. Therefore, our benchmark must evaluate whether the model can still provide textual answers about, or descriptions of, the forgotten content. To facilitate a rigorous and comprehensive evaluation, we design our QA generation process to test two distinct types of knowledge associations. The first is single-modal questions, which test the model's retention of purely textual facts, where the prompt explicitly names the concept (e.g., "What is the primary ability of \texttt{[VLUBC000]}?"). The second is multi-modal questions, which test the model's ability to visually recognize the concept in an image and link it to its factual knowledge (e.g., "What is the primary ability of the character shown in the image?"). Our generation pipeline systematically iterates over the attributes defined in each blueprint, applying pre-defined QA templates to generate corresponding single-modal and multi-modal question-answer pairs for each attribute.

\textbf{Data Filtering.} A critical final step in our pipeline is to rigorously verify that our synthetic content does not inadvertently replicate real-world intellectual property. Recent work has documented significant memorization risks in generative models; LLMs can reproduce protected textual data~\citep{carlini2021extracting, kiyomaru2024comprehensive, morris2025much}, while image generation models can replicate copyrighted visual content~\citep{somepalli2023understanding, somepalli2023diffusion, carlini2023extracting}. To mitigate these dual-modality risks within our own generation process, and to prevent our pipeline from reproducing existing protected content, we implemented a verification process for both textual and visual components. All generated attributes were cross-referenced against public trademark databases (WIPO)\footnote{https://www.wipo.int/}, and all generated reference images were analyzed using reverse image search tools\footnote{https://lens.google/} to ensure no significant visual overlap with existing copyrighted works. Any concepts that failed these checks were discarded and regenerated. The final, verified dataset, \textbf{\textsc{CoVUBench}}, consists of 2,420 VQA pairs associated with our 20 unique copyright concepts.

\subsection{Evaluation Metrics}

Our evaluation protocol is predicated on the dataset delineated in section~\ref{Dataset Construction}. Following standard practices in unlearning evaluation, we employ a two-stage pipeline~\citep{maini2024tofu, mabenchmarking, liu2025protecting}. In stage 1, a base model is fine-tuned on a training set $\mathcal{D}_{train}$ to simulate the  memorization of synthetic copyright concepts. Subsequently, a subset $\mathcal{D}_{forget} \subset \mathcal{D}_{train}$ is designated as the target for unlearning, representing content blocklisted at the request of a copyright holder. The complement set, $\mathcal{D}_{retain} = \mathcal{D}_{train} \setminus \mathcal{D}_{forget}$, constitutes the data intended to be preserved initially. In stage 2, various unlearning algorithms are applied to the fine-tuned model with the objective of removing information pertaining specifically to $\mathcal{D}_{forget}$. To assess unlearning robustness, we introduce a held-out test set, $\mathcal{D}_{test}$, featuring novel visual compositions and textual queries for the same underlying concepts as $\mathcal{D}_{forget}$, unseen during training. As evaluating utility across the entire $\mathcal{D}_{retain}$ can be computationally prohibitive, $\mathcal{D}_{retain'}$ is strategically sampled to include representative VQA pairs associated with each non-blocklisted concept  allowing for efficient utility preservation. 

Fundamentally, an effective copyright unlearning procedure must satisfy criteria aligned with two primary stakeholder perspectives: copyright content holders, focused on the efficacy of removal, and model deployers, concerned with preserving model utility and practical feasibility. Therefore, achieving truly effective unlearning hinges upon the successful reconciliation of these dual perspectives within the evaluation framework.

\noindent \textbf{Copyright Holder's Perspective.} From the copyright holder's perspective, the primary objective is the effective and robust erasure of their blocklisted content $\mathcal{D}_{forget}$. This entails preventing the model from reproducing the content, not only in its original form but also across various contexts and semantic similarities. To quantify this, we introduce three metrics.

\begin{itemize}
    \item \textbf{Efficacy:} While models may exhibit flexibility in phrasing, core information about copyrighted content often hinges on specific keywords or attributes defined in the persona blueprints. To measure the direct reproduction of these core concepts, we compute the Exact Match (EM) score on the forget set, $\text{EM}(\mathcal{D}_{forget})$. This metric calculates the average recall rate of these predefined keywords within the model's predicted answers for questions in $\mathcal{D}_{forget}$. A lower EM score indicates more effective forgetting of specific terminology associated with the blocklisted content. We report $1 - \text{EM}(\mathcal{D}_{forget})$ so that higher values indicate better forgetting.

    \item \textbf{Generality:} Copyrighted content can manifest in diverse forms, potentially deviating from the specific instances seen during training. To evaluate the robustness of forgetting against such variations—both visual and textual—we measure the EM score on the held-out test set $\mathcal{D}_{test}$, denoted $\text{EM}(\mathcal{D}_{test})$. We report $1 - \text{EM}(\mathcal{D}_{test})$, where higher values signify more robust forgetting.

    \item \textbf{Divergence:} Beyond exact keyword reproduction, a significant risk lies in the generation of near-duplicates~\citep{wei2024evaluating} that remains semantically equivalent to the blocklisted material. To quantify this semantic leakage, we compute the cosine similarity between the embeddings of the ground-truth answers and the predicted answers for queries related to $\mathcal{D}_{forget}$ using a sentence embedding model\footnote{https://huggingface.co/sentence-transformers/all-MiniLM-L6-v2}. We report this semantic dissimilarity as a normalized metric, scaled to 0-100. Higher values thus indicate a more successful erasure of the underlying semantic concepts.
\end{itemize}

\noindent \textbf{Model Deployer's Perspective.} For model deployers, unlearning must be practical, minimally impacting the model's overall capabilities while fulfilling the copyright holder's request. Desirable unlearning from this perspective involves maintaining fluency and accuracy on related, non-blocklisted content within the same domain and retaining general multimodal reasoning abilities. We assess these aspects using three metrics.

\begin{itemize}
    \item \textbf{Fluency:} The unlearning process should not degrade the model's ability to generate fluent and coherent responses for content related to, but distinct from, the blocklist. We measure this by computing the ROUGE-L recall score~\citep{lin2004rouge} between the predicted answers and ground-truth answers on a designated retain subset $\mathcal{D}_{retain'}$. 
    
    \item \textbf{Specificity:} The model must retain its factual accuracy and ability to convey specific details using appropriate keywords for non-blocklisted content. We evaluate this by measuring the Exact Match score, $\text{EM}(\mathcal{D}_{retain'})$, on the same retain subset $\mathcal{D}_{retain'}$, using the keyword recall metric defined earlier. 
    
    \item \textbf{Capability:} Unlearning should ideally be targeted, leaving the model's core multimodal reasoning and world knowledge intact. We assess the preservation of these general capabilities by evaluating the unlearned model's accuracy on established external vision-language benchmarks. Specifically, we measure performance on POPE~\citep{li2023evaluating} and MMBench~\citep{liu2024mmbench}. Average accuracy across these benchmarks serves as an indicator of retained general LVLM proficiency.

\end{itemize}

By employing this comprehensive suite of metrics, our benchmark enables a nuanced evaluation of unlearning algorithms, capturing the inherent trade-offs between forgetting efficacy (satisfying copyright holders) and utility preservation (meeting deployer needs).

\subsection{Unlearning Baselines}

To establish baseline performance on our benchmark, we evaluate a selection of representative machine unlearning algorithms, chosen to cover distinct conceptual approaches prevalent in recent literature. Each method is applied during stage 2 of our evaluation pipeline, starting from the model fine-tuned in stage 1. Let $\mathcal{L}({\mathcal{D}_{forget}}, \theta)$ denote the standard negative log-likelihood loss for a model with parameters $\theta$ on $\mathcal{D}_{forget}$. 

\noindent \textbf{Gradient Ascent (GA)~\citep{jang2023knowledge}:} This straightforward approach directly maximizes the loss on the forget set $\mathcal{D}_{forget}$ to discourage the model from generating the specific target answers associated with the blocklisted content. 

\noindent \textbf{Gradient Difference (GD)~\citep{liu2022continual}:} In this approach, a regularization term is introduced by simultaneously performing gradient descent on the retain set $\mathcal{D}_{retain}$. The objective balances maximizing the loss on $\mathcal{D}_{forget}$ with minimizing the loss on $\mathcal{D}_{retain}$, aiming to preserve performance on non-blocklisted content. 

\noindent \textbf{KL Divergence Regularization (KL)~\citep{yao2024machine}:} Instead of minimizing the loss directly on $\mathcal{D}_{retain}$, this method minimizes the Kullback-Leibler (KL) divergence between the output distribution of the unlearning model ($\theta$) and the original fine-tuned model ($\theta_{\text{orig}}$ from stage 1). This encourages the unlearning model to maintain its original behavior on $\mathcal{D}_{retain}$. 

\noindent \textbf{Direct Preference Optimization (DPO)~\citep{maini2024tofu}:} This approach reframes unlearning as a preference alignment task~\citep{rafailov2023direct}. It trains the model to prefer predefined refusal responses (e.g., "I cannot provide information on that copyrighted entity.") over the original answers for questions in $\mathcal{D}_{forget}$. An additional loss term using $\mathcal{D}_{retain}$ might be included for utility preservation.

\noindent \textbf{Negative Preference Optimization (NPO)~\citep{zhangnegative}:} NPO adapts the preference optimization framework specifically for unlearning by treating the entire forget set $\mathcal{D}_{forget}$ as negative preference data. The objective aims to decrease the likelihood of the model generating the original answers from $\mathcal{D}_{forget}$ compared to a reference model (the original fine-tuned model), without explicitly optimizing towards refusal answers during the NPO phase itself.

These selected baselines represent a spectrum of current unlearning techniques, allowing for a comprehensive assessment of their strengths and weaknesses on our proposed vision-language copyright unlearning benchmark.

\section{Experiments}

\begin{table}[t]
\centering
\renewcommand{\arraystretch}{1.2}
\setlength{\tabcolsep}{3pt}
\normalsize
\begin{tabular}{lcccc}
\toprule
Method & ROUGE & \makecell{EM\\($\mathcal{D}_{train}$)} & \makecell{EM\\($\mathcal{D}_{test}$)} & Acc. \\
\midrule
LLaVA-Phi-3B & 76.63 & 99.65 & 98.16 & 74.86 \\
LLaVA-1.5-7B & 78.15 & 99.94 & 98.95 & 73.55 \\
\bottomrule
\end{tabular}
\caption{Stage 1 fine-tuning performance for LLaVA-Phi-3B and LLaVA-1.5-7B, evaluated on ROUGE, EM ($\mathcal{D}_{train}$, $\mathcal{D}_{test}$), and external benchmark Accuracy (Acc.).}
\label{tab:stage1_performance}
\end{table}

\subsection{Experimental Setup}
We conduct our experiments using two base models: LLaVA-Phi-3B~\citep{zhu2024llava} and LLaVA-1.5-7B~\citep{liu2024improved}. For both the initial fine-tuning (stage 1) and the subsequent unlearning (stage 2), we use the AdamW optimizer and employ parameter-efficient fine-tuning via LoRA~\citep{hu2022lora}. We set the LoRA rank $r=64$ and alpha $\alpha=128$ for all experiments.Hyperparameters were set as follows: We trained LLaVA-Phi-3B for 5 epochs and LLaVA-1.5-7B for 7 epochs (this applies to both stages). The learning rate was set to $5 \times 10^{-4}$ for the stage 1 fine-tuning and $5 \times 10^{-5}$ for the stage 2 unlearning for both models. For stage 2, we designate 5\% of $\mathcal{D}_{train}$ as the forget set ($\mathcal{D}_{forget}$). 

\begin{table*}[t]
\centering
\renewcommand{\arraystretch}{0.60}
\scriptsize 
\resizebox{\textwidth}{!}{
\begin{tabular}{lcccccc}
\toprule
\multicolumn{7}{c}{\textbf{LLaVA-Phi-3B}} \\
\midrule
Method & Efficacy & Generality & Divergence & Fluency & Specificity & Capability \\
\midrule
GA  & 47.63 & 56.33 & 83.57 & 73.00    & 99.33 & 74.86  \\
GD  & 45.96 & 53.83 & 83.34 & 74.78 & 99.49 & 74.36 \\
KL  & 46.79 & 57.33 & 83.72 & 73.03 & 99.33 & 74.80 \\
DPO & 23.54 & 31.09 & 97.17 & 74.92 & 100.00   & 74.77  \\
NPO & 48.25 & 51.50  & 83.72 & 74.82 & 99.75 & 74.85 \\
\midrule
\multicolumn{7}{c}{\textbf{LLaVA-1.5-7B}} \\
\midrule
Method & Efficacy & Generality & Divergence & Fluency & Specificity & Capability \\
\midrule
GA  & 82.09 & 86.36 & 71.02  & 50.54 & 93.60   & 74.51 \\
GD  & 7.50    & 10.69  & 94.46 & 53.79  & 98.49 & 78.16  \\
KL  & 72.09 & 73.07 & 75.20   & 51.56 & 95.12  & 74.80   \\
DPO & 11.25  & 10.15  & 92.55 & 72.02 & 98.48  & 79.52 \\
NPO & 42.50   & 39.23 & 90.60 & 60.90 & 98.07 & 78.29 \\
\bottomrule
\end{tabular}
}
\caption{Stage 2 unlearning results on LLaVA-Phi-3B and LLaVA-1.5-7B across our six proposed metrics. The results highlight the trade-off between forgetting (Efficacy, Generality, Divergence) and utility (Fluency, Specificity, Capability).}
\label{tab:unlearning_llava_metrics}
\end{table*}

\subsection{Experimental Results}
\textbf{Stage 1: Base Model Fine-tuning Performance.} We first evaluate the base models after stage 1 fine-tuning, reporting ROUGE, Exact Match (EM) on the train ($\mathcal{D}_{train}$) and test ($\mathcal{D}_{test}$) sets, and average accuracy (Acc.) on external benchmarks (POPE, MMBench). As shown in Table~\ref{tab:stage1_performance}, both LLaVA-Phi-3B and LLaVA-1.5-7B achieve high ROUGE scores indicating high fluency, and near-perfect EM scores on $\mathcal{D}_{train}$, confirming effective memorization of core concepts. This performance generalizes well to the held-out $\mathcal{D}_{test}$ while maintaining solid accuracy on external VLM benchmarks. These results confirm that the models successfully learned the synthetic copyright concepts, providing a robust foundation for evaluating unlearning algorithms in stage 2.

\textbf{Stage 2: Unlearning Algorithm Comparison.} Table~\ref{tab:unlearning_llava_metrics} presents the main results for the five unlearning baselines, revealing a clear trade-off between the two stakeholder perspectives, especially on the larger LLaVA-1.5-7B model. From the copyright holder's perspective, standard gradient-based methods (GA, KL) are the most effective. On the 7B model, GA achieves the highest Efficacy and Generality, indicating robust erasure. However, this success comes at a significant cost to the model deployer, causing a catastrophic drop in Fluency to around 50 and a loss of Specificity. Conversely, preference-based methods (DPO, NPO) and the GD excel at preserving utility; DPO, for example, maintains a high Fluency (72.02) and Capability. Yet, these utility-preserving methods fail at true erasure, scoring exceptionally low on Efficacy and Generality. Notably, their high Divergence scores (frequently above 90) suggest they do not truly forget the concept but instead learn a superficial refusal strategy. The smaller LLaVA-Phi-3B model shows a similar but less pronounced trend, where utility is generally well-preserved by all methods. Overall, our results demonstrate that no current method ideally satisfies both stakeholders on the larger model; GA and KL prioritize erasure for the holder, while DPO, NPO, and GD prioritize utility for the deployer.

\subsection{Ablation Study}

\begin{figure}[t]
\centering
\includegraphics[width=\columnwidth]{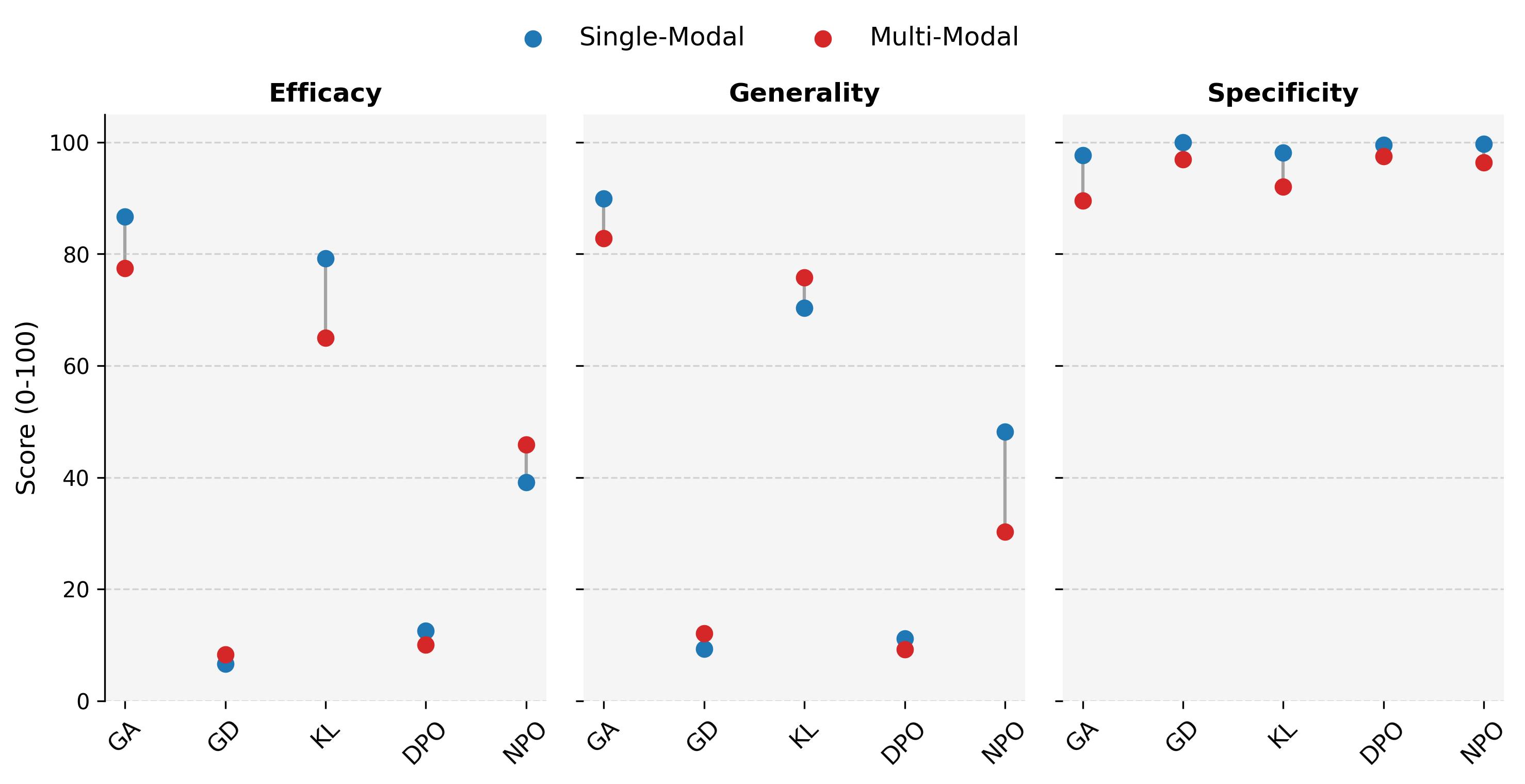} 
\caption{A comparison of single-modal (text-only) and multi-modal (vision-inclusive) performance for five unlearning baselines, evaluated across key unlearning metrics.}
\label{fig:modal_question}
\end{figure}

\textbf{Analysis on the Modality Gap in Unlearning.} Figure~\ref{fig:modal_question} reveals a consistent performance gap between single-modal (text-only) and multi-modal (vision-inclusive) queries. This "modality gap" is most pronounced in GA and KL, which achieve high single-modal Efficacy but suffer a substantial performance drop when faced with multi-modal queries that test the vision-knowledge link. This delta extends to utility, as these same methods show a greater drop in Specificity on multi-modal retain questions, indicating more collateral damage to visual reasoning. Conversely, GD, and DPO display a negligible gap, not from robust cross-modal unlearning, but because they fundamentally fail at erasure (Efficacy/Generality near-zero) in both modalities. This gap underscores that current methods struggle with true multi-modal concept erasure, likely only severing textual associations rather than disassociating the underlying visual concept.

\begin{figure*}[t]
\centering
\includegraphics[width=0.95\textwidth]{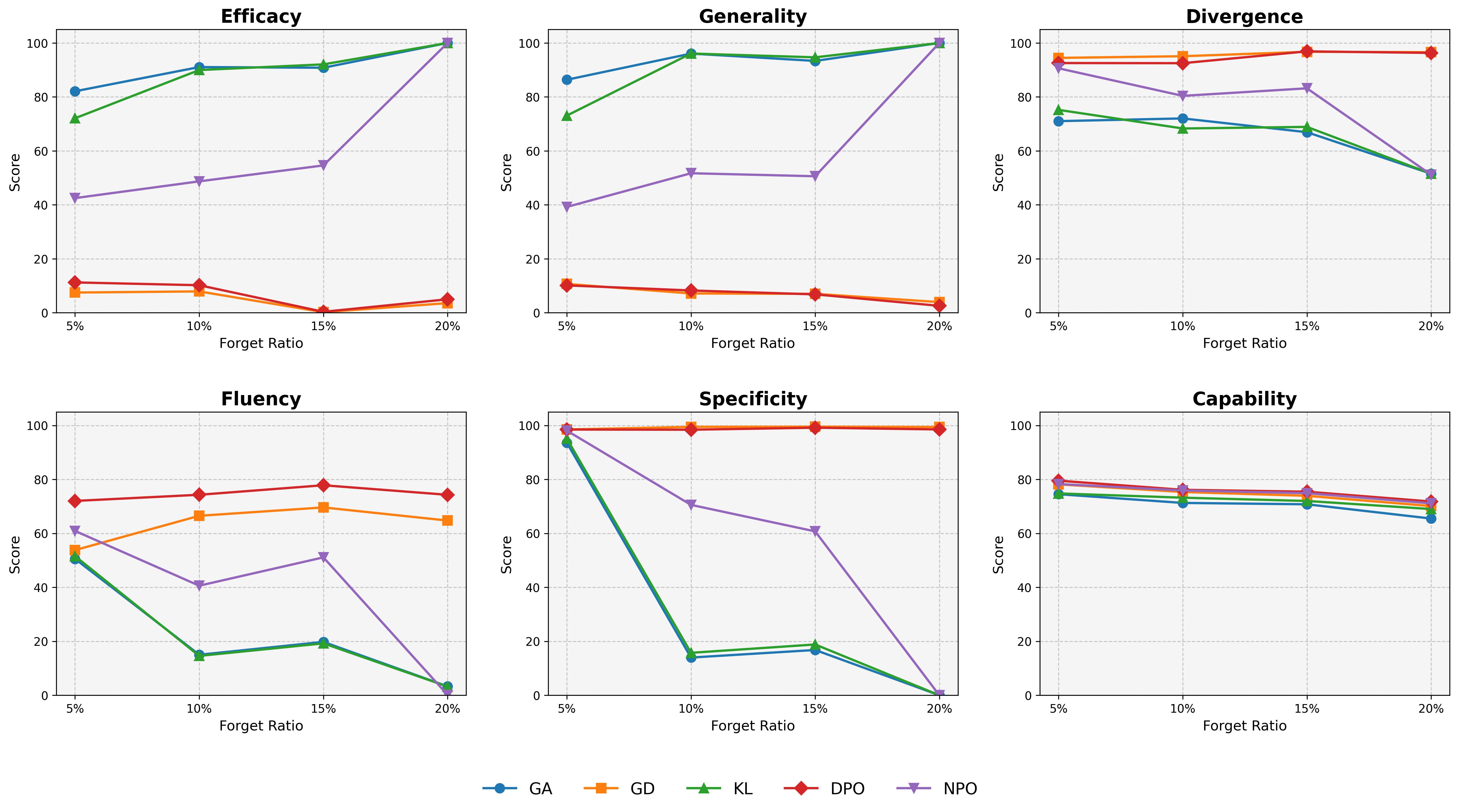} 
\caption{The impact of increasing the forget ratio (5\% to 20\%) on the performance of five unlearning baselines, evaluated across our six proposed metrics.}
\label{fig:forget_ratio_trends}
\end{figure*}

\textbf{Analysis on the Impact of Forget Ratio.} We analyze the performance of unlearning methods as the forget ratio increases from 5\% to 20\%, with results shown in Figure~\ref{fig:forget_ratio_trends}. The data reveals two distinct and conflicting behaviors: gradient-based methods (GA, KL) and NPO demonstrate a "catastrophic over-forgetting" phenomenon. As the forget ratio increases, their Efficacy and Generality improve, reaching perfect erasure at 20\%. However, this comes at a severe cost to utility, with Fluency and Specificity plummeting to 0, indicating the model's in-domain capabilities are completely destroyed. In stark contrast, GD and DPO are exceptionally robust to the increasing ratio, maintaining high and stable utility across all metrics, particularly Fluency and Specificity. Yet, these methods are entirely ineffective at the core task of erasure, with Efficacy and Generality scores remaining near-zero (often below 10) regardless of the ratio. Their consistently high Divergence scores confirm they do not truly erase the concept but instead learn a robust, superficial refusal strategy. Our findings show that no current method is both effective and robust to increasing forget ratios; GA, KL, and NPO prioritize erasure by destroying the model, while GD and DPO prioritize utility by failing to erase.

\begin{figure}[t]
\centering
\includegraphics[width=\columnwidth]{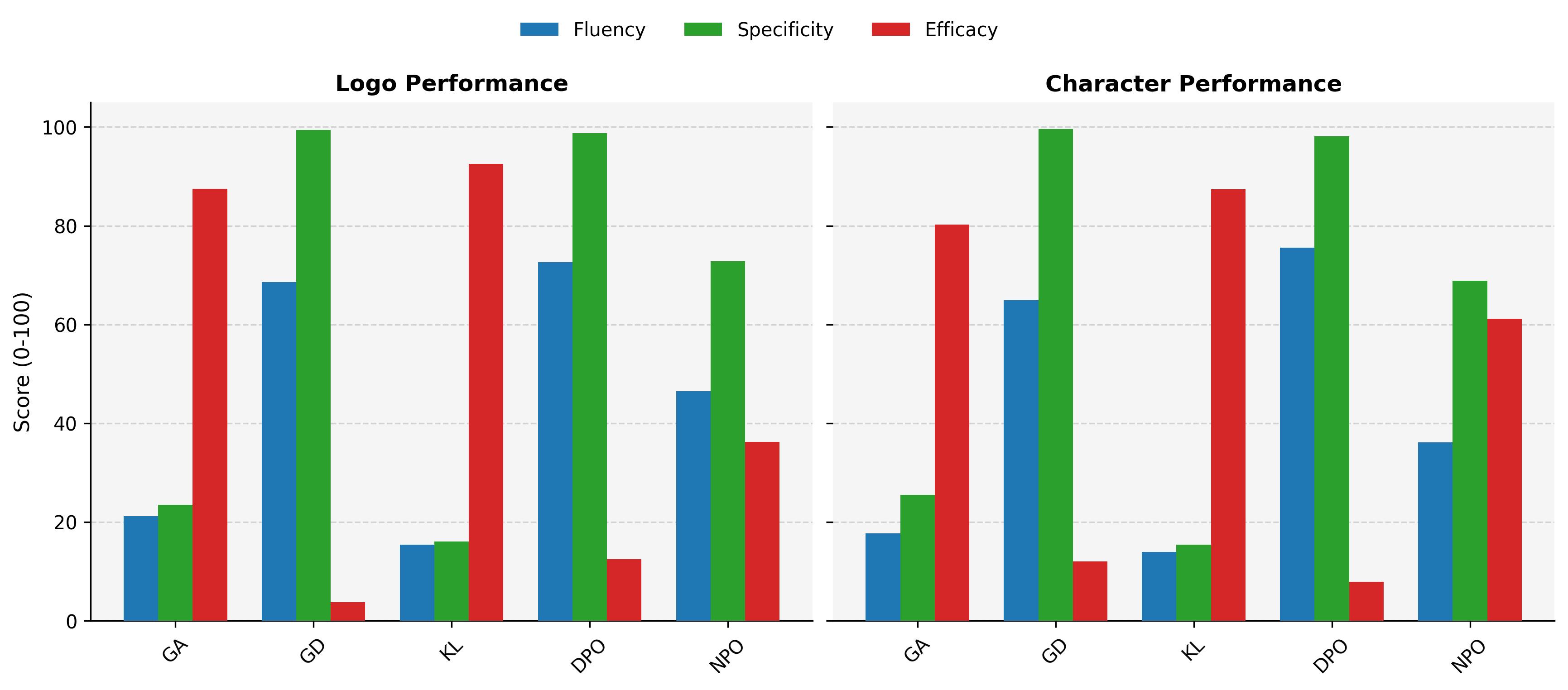} 
\caption{Performance comparison across 'logo' and 'character' domains for five unlearning baselines, evaluated on Fluency, Specificity and Efficacy.}
\label{fig:logo_character}
\end{figure}

\textbf{Analysis on Domain-Specific Unlearning.} Figure~\ref{fig:logo_character} breaks down performance by copyright domain. While the general trade-off pattern persists, the results reveal domain-specific sensitivities. GA and KL are marginally more effective at erasing logos than characters (e.g., KL Efficacy: 92.5 vs 87.4), though both domains suffer a catastrophic utility collapse. The most significant divergence is seen with NPO, which achieves substantially higher Efficacy on characters than on logos, a gap of 24.92\%p.This improved character erasure, however, is coupled with a more severe drop in Fluency. GD and DPO remain consistently ineffective at erasure, failing to forget either domain. We conjecture that NPO more effectively suppresses the complex textual 'lore' of characters than the atomic visual-name link of logos, but this broader suppression of 'lore' causes collateral damage to general fluency.

\section{Conclusion}
We introduced \textbf{\textsc{CoVUBench}}, the first dedicated benchmark for evaluating multimodal copyright unlearning in LVLMs. Our framework enables a robust evaluation by introducing a visually diverse corpus spanning multiple domains and a diagnostic VQA set designed to probe both textual-level associations and the deeper vision-knowledge link. Our stakeholder-centric evaluation, focused on the dual needs of content removal and utility preservation, demonstrated that current unlearning algorithms are fundamentally inadequate for this task. We identified a stark polarization where current methods either achieve effective unlearning by catastrophically collapsing model utility, or preserve utility by completely failing to erase the blocklisted concept. Furthermore, our analyses exposed a critical "modality gap," revealing that all methods struggle to sever the underlying vision-knowledge link even when textual associations are removed.  Our work underscores the clear and urgent need for the development of novel, specialized algorithms designed explicitly for multimodal unlearning.

\section{Acknowledgments}

This work was supported by the Institute of Information \& Communications Technology Planning \& Evaluation (IITP) grant funded by the Korea government (MSIT) [RS-2021-II211341, Artificial Intelligence Graduate School Program (Chung-Ang
University)] and by the National Research Foundation of Korea (NRF) grant funded by the Korea
government (MSIT) (RS-2025-00556246).

\section{Bibliographical References}\label{sec:reference}

\bibliographystyle{lrec2026-natbib}
\bibliography{lrec2026-example}

\end{document}